\documentclass[letterpaper, 10 pt, conference]{ieeeconf}  
\usepackage[T1]{fontenc}
\usepackage{cite}
\usepackage{svg}

\IEEEoverridecommandlockouts                              

\usepackage{graphics} 
\usepackage{epsfig} 
\usepackage{mathptmx} 
\usepackage{times} 
\usepackage{amsmath} 
\usepackage{amssymb}  
\usepackage{tikz}
\newcommand{\YY}[1]{\textcolor{black}{#1}}
\newcommand{\RBM}[1]{\textcolor{black}{#1}}
\newcommand{\ZC}[1]{\textcolor{black}{#1}}

\title{\LARGE \bf
TeleopLab: Accessible and Intuitive Teleoperation of a Robotic Manipulator for Remote Labs*
}

\author{Ziling Chen$^{1}$, Yeo Jung Yoon$^{1}$, Rolando Bautista-Montesano$^{1}$, Zhen Zhao$^{1}$, Ajay Mandlekar$^{2}$ and John Liu$^{1}$
\thanks{*This work was supported by U.S. Economic Development Administration and Industrial Base of Analysis and Sustainment}
\thanks{$^{1}$Learning Engineering and Practice (LEAP) Group, Massachusetts Institute of Technology
        {\tt\small johnhliu@mit.edu}}%
\thanks{$^{2}$Nvidia, USA
       }%
}

\begin{document}

\bibliographystyle{ieeetr}

\maketitle
\thispagestyle{empty}
\pagestyle{plain}

\begin{abstract}

Teleoperation offers a promising solution for enabling hands-on learning in remote education, particularly in environments requiring interaction with real-world equipment. However, such remote experiences can be costly or non-intuitive. To address these challenges, we present TeleopLab, a \RBM{mobile device}
teleoperation system that allows students to control a robotic arm and operate lab equipment. TeleopLab comprises a robotic arm, adaptive gripper, cameras, lab equipment \RBM{for a diverse range of applications}, a user interface accessible through smartphones, and \RBM{video call software}. 
We conducted a user study, focusing on task performance, students' perspectives toward the system, usability, and workload assessment. Our results demonstrate a 46.1\% reduction in task completion time as users gained familiarity with the system. Quantitative feedback highlighted improvements in students' perspectives after using the system, while NASA TLX and SUS \RBM{assessments} indicated a manageable workload of \RBM{38.2} and a positive usability of \RBM{73.8}. TeleopLab successfully bridges the gap between physical labs and remote education, offering a scalable and effective platform for remote STEM learning. \ZC{The project will be open-source. We will try to align the release schedule upon acceptance of this} \RBM{manuscript.}

\end{abstract}

\section{INTRODUCTION}

Remote learning, also known as distance education or online learning, refers to an educational process where students and instructors are not physically present in a traditional classroom environment. Instead, instruction and learning occur through digital platforms, enabling students to access course materials, interact with instructors, and participate in discussions from virtually anywhere \cite{Turoff1986}. The popularity of remote learning has surged in recent years, largely due to technological advancements, increased internet accessibility, and the flexibility it offers to both learners and educators \cite{Calvo2010, Hunter2015}. The global pandemic further accelerated its adoption as educational institutions sought alternatives to in-person learning \cite{Ali2020, Ma2006, Ndibalema2022, Schwartz2020}. Today, remote learning is favored for its ability to accommodate diverse learning needs, provide access to a wider range of courses, and support continuous education despite geographical constraints \cite{Ciloglu2023, Fitter2020, Gao2023, MacNeill2024, Susanti2023, Pokorný2023}.

Despite the widespread adoption of remote learning, it presents significant challenges, particularly in delivering hands-on experiences essential for disciplines that require practical skills \RBM{and equipment} \cite{Ma2006, May2023}. Traditional in-person education enables direct interaction with physical equipment, allowing students to develop \RBM{sensorimotor} skills, troubleshoot in real time, and gain a deeper understanding of complex systems through observation and manipulation. In contrast, remote learning often struggles to replicate these experiences effectively \cite{Han2021, Ismawati2023}. The absence of physical interaction can limit the ability of students to interact with the material intuitively, leading to gaps in learning outcomes \cite{Ackovska2017, Chang2023, Dhanapal2014, Filgona2016, Holstermann2010}.  
\begin{figure}[htbp]
      \centering
      \makebox{\includegraphics[width = 3.3in]{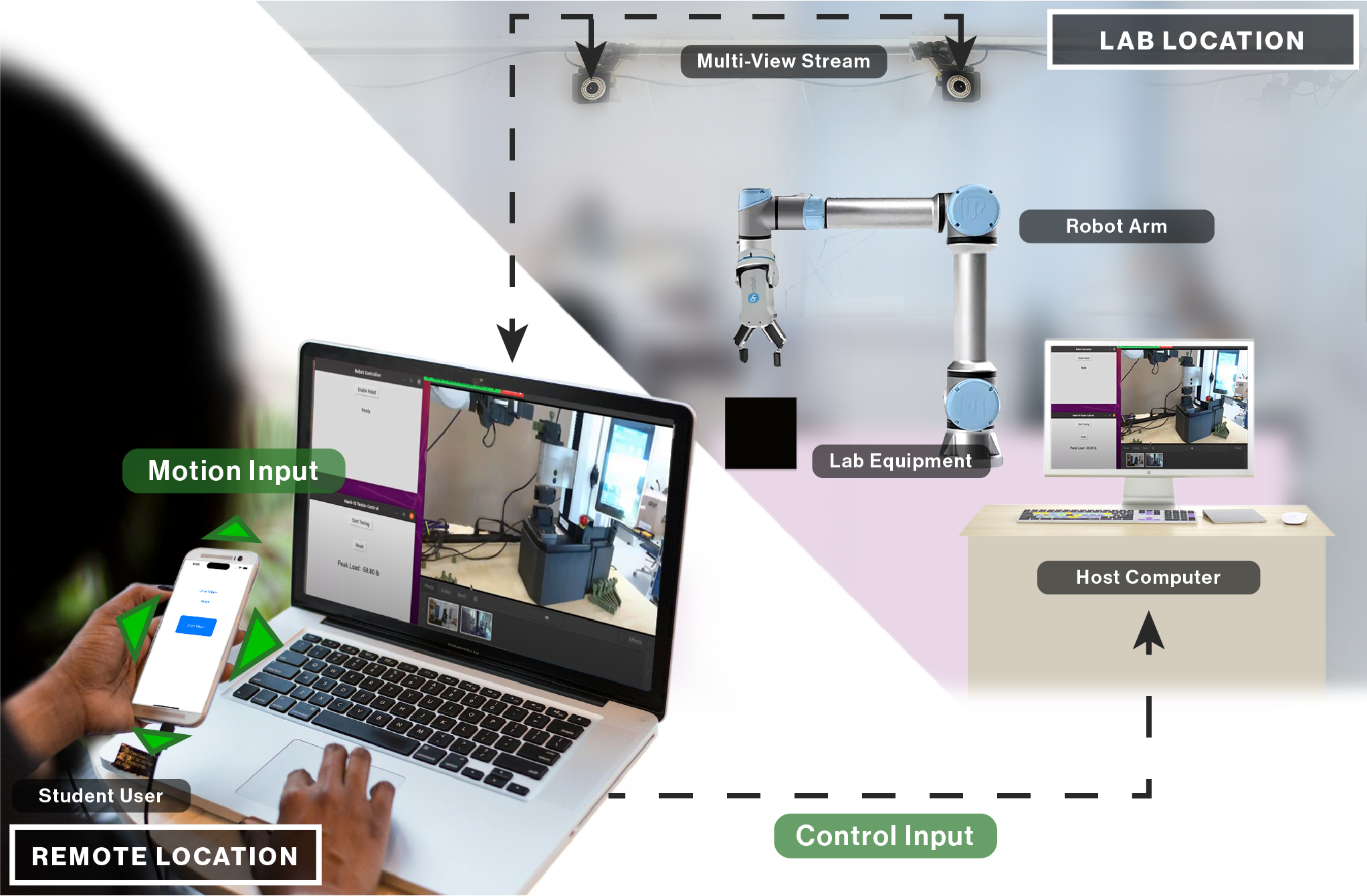}}
      \caption{\textbf{TeleopLab System Diagram.} The user controls the robot by moving their phone to provide waypoints \RBM{that are transformed to commands to manipulate} lab equipment at a remote location. The camera views are streamed to a host computer and viewed on the remote desktop.}
      \label{figure1}
   \end{figure}

Teleoperation allows students to remotely control real equipment, offering a more authentic hands-on experience \cite{Kaarlela2022, Kondratyev2022, Rakshit2017, Tan2019}. Unlike simulations, which cannot capture the complexities of real-world systems \cite{Davis2024, McAdams2024, Kortemeyer2024, Reiners2014, Farley2016}, teleoperation enables students to interact \RBM{with physical phenomena in real contexts}. Teleoperation in education has traditionally relied on keyboards \cite{Kondratyev2022, Tan2019} and mice \cite{Kaarlela2022, Tan2019}, which are often \RBM{not} intuitive and limited to basic robotics tasks. More intuitive systems, such as those that use haptic devices \cite{Esfandiari2023, Yilmaz2024}, cameras, wearable devices \cite{Yang2023, Zhu2022} and virtual reality (VR) joysticks \cite{Tan2019, livatino2021intuitive, Dengxiong2024, Rakshit2017}, offer precise control, but face accessibility and scalability issues due to high costs \cite{ashtari2020creating, ardiny2018role, Kavanagh2017} and specialized training needs \cite{Fernandez2017, Kavanagh2017, Mora2017}, limiting their use in education. 

In response to the challenges of teleoperation in remote learning, we present TeleopLab, a teleoperation system that enables students to engage in hands-on activities \YY{with a robotic arm using their smartphones (Fig. \ref{figure1}).} Our solution addresses the key limitations of existing teleoperation systems by offering an intuitive and accessible interface that reduces the need for expensive and specialized input devices such as VR sets and haptic devices. With TeleopLab, students can use both Android phones and iPhones to capture 
motion and operate a robotic arm remotely, providing a seamless and immersive learning experience. 
Our contributions are fourfold:
\begin{enumerate}
  \item We present the TeleopLab platform, which harnesses the ubiquity of smartphones to provide an accessible and scalable teleoperation solution \YY{for robotic arm manipulation}. TeleopLab offers a more natural and engaging way for students to interact with remote equipment, enhancing their learning experience.
  \item We improve the user experience by addressing latency, intuitiveness, and compatibility. 
  The system was designed \RBM{considering students as the target audience.} It minimizes distraction and cognitive load. 
  \item We compare different remote learning options and demonstrate that teleoperation, particularly through our system, offers the best cost-effectiveness to enable lab activities remotely in a case study.
\end{enumerate}

\section{Related Work}

\textbf{Current Teleoperation Methods in Education:} In educational contexts, teleoperation methods have predominantly relied on traditional input devices such as keyboards \cite{cabrera2024physical, Tan2019}, and mouse controls \cite{pleul2011tele, hincapie2013implementation, terkowsky2010developing, moreno2022teleoperated, Kaarlela2022, Kondratyev2022, Tan2019}, particularly in robotics teaching. These approaches allow students to manipulate robots or systems remotely, focusing primarily on executing predefined commands and sequences. Most systems focus on teaching specific robotic functions, such as navigation \cite{cabrera2024physical} or robot control \cite{Kondratyev2022, saku2023, Kaarlela2022}. Some extend teleoperation to broader fields that could benefit from robotic interaction, including material science \cite{pleul2011tele, ortelt2014development}, manufacturing \cite{terkowsky2010developing, hincapie2013implementation}, chemistry \cite{moreno2022teleoperated} and mechatronics \cite{lopez2022adapting}. However, the use of such input methods is often unintuitive and disconnected from the natural hand movements or gestures that would be used in real-world manipulation \cite{Mandlekar2018}. This creates a disconnection between the learning experience and real-life applications, resulting in a steeper learning curve and reduced engagement.

\textbf{Current Intuitive Teleoperation Systems:} More intuitive teleoperation systems have emerged in recent years, leveraging specialized input devices such as haptic devices \cite{gonzalez2021advanced, tavakoli2008haptics, Esfandiari2023, Yilmaz2024, Rakshit2017}, VR joysticks \cite{Tan2019, livatino2021intuitive, Dengxiong2024, Rakshit2017}, and wearable devices \cite{Yang2023, Zhu2022}. These devices provide users with precise control over robotic systems, making them highly effective in applications requiring accurate manipulation, such as surgical \cite{tavakoli2008haptics, Esfandiari2023, Yilmaz2024} or industrial automation \cite{gonzalez2021advanced}. Their potential for remote learning, particularly in STEM education, is significant, as they allow students to interact with robots in a manner that closely mimics real-world actions. However, the scalability of these systems is hindered by the high cost of maintaining the software/hardware \cite{ashtari2020creating} and distribution \cite{ardiny2018role, Kavanagh2017}, and the significant training required for effective use \cite{Fernandez2017, Kavanagh2017, Mora2017}. Therefore, while specialized input devices enhance the realism of teleoperation, their lack of scalability presents a barrier to integrating them into broader educational practices.

\textbf{Phone-based Teleoperation:}  
Recently, phone-based teleoperation has gained traction, particularly in crowdsourcing applications, due to the ubiquity of smartphones and advancements in communication technologies. Projects like RoboTurk \cite{Mandlekar2018} and TeleMoMa \cite{Dass2024} have showcased the potential of phone-based teleoperation for data collection. Still, this focus on generating datasets for machine learning differs from the needs of an educational setting. In education, the system should prioritize real-time feedback \cite{Kaarlela2022, david2024}, ease of use \cite{saku2023}, and hands-on engagement \cite{kang2010quality}. \YY{In this study, we focus on providing students with an interactive and intuitive experience with real-world equipment} \RBM{through diverse tasks that include moving, grabbing, and pushing in multiple robot arm platforms.} 

\section{TeleopLab System Design}
\subsection{System Design} 
TeleopLab is composed of two primary components: the teleoperation station and the user endpoint. The teleoperation station includes a robotic arm, a gripper, multiple cameras, and a host computer (server) for processing input commands and controlling the robot. On the user end, a student needs a smartphone with a custom app and a remote monitoring platform such as Zoom. 

\textbf{Robotic Arm.} 
\YY{TeleopLab platform is designed to be adaptable to diverse educational and industrial environments, supporting various robotic arms. The control system is built on Robot Operating System (ROS) Noetic, running on Ubuntu 20.04, enabling compatibility with a wide range of ROS-supported robotic manipulators. For this study, we implemented our system on both an ABB IRB-120 and a UR5e robotic arm, demonstrating its adaptability across different hardware configurations. }
Although ABB provides support for the robotic arm via the ABB Driver within the ROS Industrial package, this driver relies on the TCP-based Robot Web Service (RWS), which was designed for general communication purposes and operates at a lower frequency. This setting is less suitable for high-frequency, real-time teleoperation. \YY{To achieve smoother, more responsive teleoperation, we implemented a Externally Guided Motion (EGM) control system based on Google Protocol Buffers over UDP. EGM leverages a higher update rate for direct motion control, resulting in less latency and more fluid robotic arm movement.} \RBM{Plus, EGM allows to remotely configure  most of the robotic arm's settings available in the host machine.} 
This setup minimizes 
onsite staff intervention, making it easier to deploy the system in educational environments where technical support may be limited.

\textbf{End-Effector.} 
\YY{Users can install different end-effectors depending on task requirements and their specific needs. For our study, we used the InstaGrasp \cite{Zhou2023} and a Robotiq two-finger gripper. InstaGrasp is a 3D-printed adaptive gripper that offers a cost-effective and modular solution for remote manipulation tasks.}


\textbf{Vision System.} \YY{Two cameras were used to provide multiple viewing angles during teleoperation.} The first is a RealSense  D405 3D camera mounted on the end effector of the robotic arm, providing a direct view of the manipulation tasks. The second camera, a Logitech C920x HD Pro Webcam, is positioned on a tripod to offer a wide-angle view of the entire workspace, \RBM{providing} 
students with an overview of the environment. Both camera feeds are \RBM{live} streamed using Cheese, a native camera software for Ubuntu, allowing seamless integration with the system. Students can switch between these two camera views based on their preferences once they join the Zoom session and gain remote control of the system. 

\textbf{Server.} An HP EliteBook 840 laptop (i5-4300U @1.9GHz, 8GB RAM) was selected as the host computer for TeleopLab. 
The laptop runs Ubuntu 20.04 and ROS Noetic, serving as the core system for managing user input and robot control. Leveraging TeleMoMa \cite{Dass2024}, we developed interfaces for user commands, the robot, and lab equipment, providing students with a seamless way to control the robotic arm and interact with lab tools remotely.

The user command interface uses a TCP-based communication structure to connect the student’s smartphone to the server, processing and relaying real-time inputs to the robot. While TeleMoMa \cite{Dass2024} supports remote teleoperation, it was not optimized for long-distance use, so we implemented port forwarding for enhanced security, rerouting the public IP at the lab location to the host computer and avoiding direct IP assignment. This setup enables easy and secure connections for students.

For the robot interface, the system performs a coordinate transformation based on the user's input to align the user's perspective with that of the robotic arm. After this transformation, the system utilizes TracIK \cite{Beeson2015} as the inverse kinematics (IK) solver to convert the user's input pose into joint space. TracIK offers robust and efficient solve rates, with an average solve time of less than 1 millisecond. In addition, we extended the TeleMoMa robot interface by developing a \RBM{user interface (UI)} that allows students to connect their phones, restore lost connections, and enable the robotic arm with ease.

For the lab equipment interface, we implemented a serial communication-based server to control the Mark-10 F305-EM tensile tester. The tensile tester is fully controlled by the host computer through the PC Control function. To facilitate remote operation, we created a simple, clickable application with a user-friendly UI containing two buttons: one to start testing and one to reset the tester (Fig. \ref{figure1}). This interface enables students to perform material strength tests remotely with minimal technical overhead, allowing for a seamless hands-on learning experience. 
 
\textbf{User Endpoint.} The user endpoint provides students with two primary interfaces for interacting with the TeleopLab system: a smartphone app for controlling the robotic arm through motion sensing and a remote monitoring platform for accessing the desktop of the host computer and viewing the camera feeds of the workspace \RBM{(Zoom)}. 

We developed the app for both iOS and Android platforms to ensure accessibility across a wide range of devices. For iOS (and iPads), the app uses ARKit from Apple, which leverages real-time pose estimation by utilizing both the camera frames and the onboard motion sensor data of the phone. Similarly, the Android version employs Google ARCore. 

To streamline operations, we implemented relative pose estimation, which 
\RBM{sets the current} pose as the origin when the "start teleop" button is pressed. All subsequent movements are estimated relative to this origin until the button is released. This approach eliminates the need 
to manage absolute positions during complex or long operations, enhancing usability and reducing complexity. 

The connection to the server is automatically established when the app starts, with the IP address and port hard-coded into the app. The app is distributed through TestFlight for iOS and by manually emailing the APK file for Android, making it easily accessible for students.

\textbf{Latency Reduction.} Minimizing latency in TeleopLab is crucial for maintaining a smooth learning experience. Unlike teleoperation systems for data collection, where users can adapt to delays, students need minimal latency to stay focused on tasks. Our study broke down latency into three main components: internet delay, inverse kinematics (IK) computation, and motion execution speed. While internet delay was out of our control, IK computation had a negligible impact, with solve times under 1 millisecond.

Our primary focus was on reducing motion execution speed, where we successfully decreased latency from 500 milliseconds to 10 milliseconds. By increasing joint speed to 100 degrees per second, we improved responsiveness. To ensure smooth movements, we applied a low-pass filter to remove data jittering and implemented a solution comparison process to prevent singularities or physical joint limits, ensuring stable, real-time control during student operations.
\begin{figure*}[htbp]
    \centering
    \includegraphics[width=0.90\textwidth]{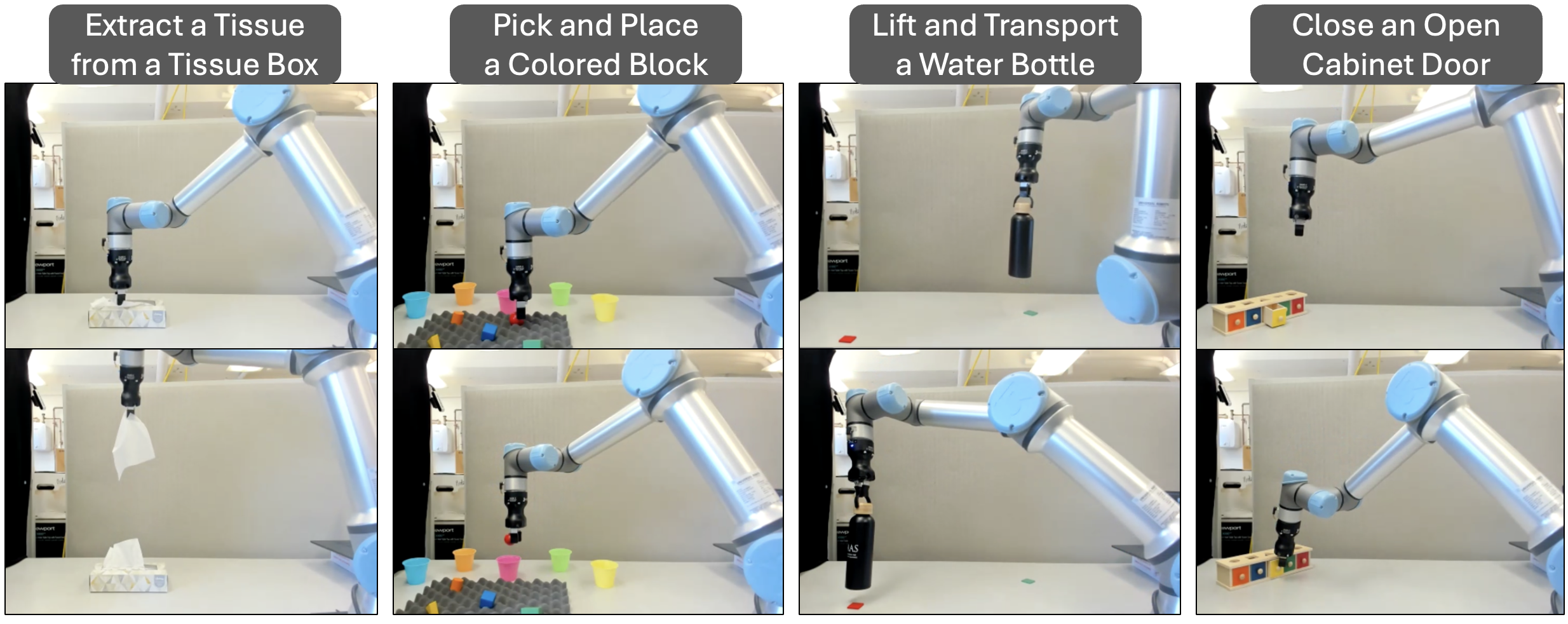}
    \caption{\textbf{TeleopLab for Different Tasks.} Our TeleopLab system is designed for users to carry out a variety of object manipulation tasks (here shown: a UR5e collaborative robot and a Robotiq two-finger gripper.) }
    \label{figure2}
\end{figure*} 
\textbf{Intuitiveness.} Since TeleopLab is designed for students with little experience with using robotics, \RBM{making} system 
intuitiveness 
a key priority. 
\RBM{Mapping the degrees of freedom between the student’s hand and the robot arm represented a challenge, as it had to be straightforward and easy to understand}. 
To address this, we used an approach similar to the Geometry Approach \cite{Lee1984} in inverse kinematics (IK), which differentiates between various configurations (e.g., left and above arm vs. left and below arm). Students expect the robotic arm to maintain a consistent configuration throughout, but IK solvers can sometimes generate inconsistent solutions. To improve predictability, we fixed the configuration indicator or reduced the DOF control when possible, making the system more user-friendly.

Additionally, we introduced virtual fences around equipment to act as both a safety barrier and a guide. For example, when loading a sample into the tensile tester, the fence stops the robotic arm from advancing beyond a certain point, ensuring safe interaction and providing clear visual feedback to the user.

\YY{By following the design principles, we conducted several tests to demonstrate TeleopLab’s versatility in real-world scenarios. Fig. \ref{figure2} illustrates how these tests were carried out using a UR5e collaborative robot equipped with a Robotiq two-finger gripper.}

\subsection{Case Study: Remote Lab Solution Comparison} The Berkshire Innovation Center (BIC) offers the Berkshire Manufacturing Academy (BMA) \RBM{program}. \RBM{It is} designed to bridge the gap between local supply chain capabilities and the needs of larger manufacturers. A critical part of the BMA curriculum involves hands-on lab activities, where students design, print, and measure parts to understand manufacturing variations using the DMAIC (Define, Measure, Analyze, Improve, Control) framework. One essential lab task requires students to use a tensile strength tester to measure the yield load of printed parts, providing direct engagement with the physical properties of materials.
\begin{table*}
\centering
\caption{Pugh Chart on Comparison of Different Remote Lab Solutions.}
\label{pughChart}
\begin{tabular}{|l|c|c|c|c|c|c|}
\hline
\textbf{}           & \textbf{Mailing Kits} & \textbf{Simulation (PC)} & \textbf{Simulation (VR)} & \textbf{TeleMoMa (keyboard)} & \textbf{TeleMoMa (phone)} & \textbf{TeleopLab} \\ \hline
\textbf{Accessibility}   & 0                   & 1                      & -2                    & 1                         & 1                     & 1          \\ 
\textbf{Engagement}      & 0                   & -2                     & -1                    & -1                        & 0                     & 0          \\ 
\textbf{Flexibility}     & 0                   & 0                      & 0                     & 1                         & 1                     & 1          \\ 
\textbf{Intuitiveness}   & 0                   & 0                      & 0                     & -1                        & -1                    & 0          \\
\textbf{Interactivity}  & 0                   & -1                     & -1                    & -1                        & -1                    & -1         \\ 
\textbf{Realism}         & 0                   & -2                     & -2                    & 0                         & 0                     & 0          \\ 
\textbf{Safety}          & 0                   & 2                      & 1                     & 1                         & 1                     & 2          \\ 
\textbf{Scalability}     & 0                   & 2                      & 2                     & 1                         & 1                     & 1          \\ \hline
\textbf{Total}           & 0                   & 0                      & -3                    & 1                         & 2                     & 4          \\ \hline
\end{tabular}

\end{table*}
With the shift to online learning, maintaining the hands-on component of such activities presented significant challenges. Tasks like tensile strength testing require multiple iterations of measurement and parameter adjustments, which would necessitate numerous shipments of parts and equipment between students and the academy, an impractical and unsustainable approach, especially with growing class sizes. Although simulations offer scalability, they fall short in capturing real-world nuances like sample positioning during testing, which can greatly affect measurement accuracy. Moreover, simulations lack the flexibility to easily adapt to changing educational objectives, requiring costly and time-consuming adjustments.

To evaluate potential solutions, a Pugh chart (Table \ref{pughChart}) was developed based on the qualitative feedback from preliminary testing with a simple tower de-stacking task, comparing two types of TeleMoMa interfaces (keyboard and phone) and TeleopLab (our solution). Simulations were excluded from this testing phase. The chart revealed that teleoperation via phone and our TeleopLab system outperformed other approaches in scalability, accessibility, and safety. While both types of teleoperation improved interactivity compared to physical kits, TeleopLab demonstrated better safety for surrounding equipment and intuitiveness, offering real-time control and adjustments. This allows students to replicate the hands-on experience remotely, controlling a robotic arm via smartphones to interact with real-world equipment as the tensile strength tester. By providing a flexible, scalable, and intuitive platform, TeleopLab ensures that remote learning can still deliver the practical experience essential for manufacturing education.

\section{Data Collection}
\subsection{Task Design}
We modified the in-person activity to fit the online format while preserving as much hands-on interaction as possible. The redesigned task ensures student interaction with physical equipment via teleoperation, maintaining the practical aspects of the learning experience. The task is structured as follows: 

\begin{enumerate}
    \item \textbf{Connect to the server:} Students independently connect to the TeleopLab server using the app on their smartphone.
    \item \textbf{Control the robotic arm:}
    \begin{enumerate}
        \item \textbf{Pick up a specimen:} Once connected, students remotely control the robotic arm to pick up a specimen from a designated reloading location.
        \item \textbf{Place the specimen on the tensile tester:} Students then guide the robotic arm to place the specimen onto the Mark-10 tensile tester for measurement.
        \item \textbf{Control the testing process:} Students operate the tester remotely, starting the test, recording the results, and resetting the tester for the next specimen.
        \item \textbf{Repeat the process:} Students return to the reloading location to pick up a new specimen and repeat steps a-d.
    \end{enumerate}
    
\end{enumerate}

Although automatic preloading using a stack of specimens and automatic unloading with a linkage mechanism were designed, they were not implemented during the data collection phase due to time constraints. As a result, onsite staff assisted with preloading and unloading the tested specimens. Despite this, the task design ensures that students control most aspects of the process, closely replicating the original in-person activity.

\subsection{Data Collection}
\begin{figure}[htbp]
      \centering
      \makebox{\includegraphics[width = 2.8in]{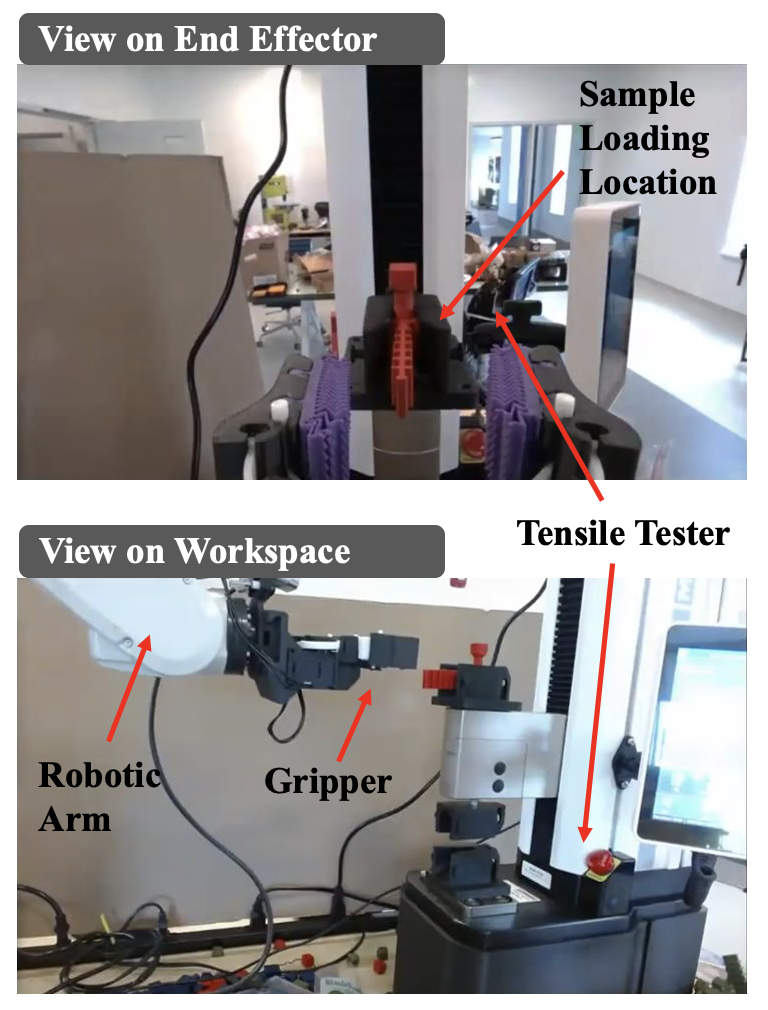}}
      \caption{\textbf{Remote Tensile Strength Testing} }
      \label{figure3}
   \end{figure}

\YY{We conducted our pilot test in an educational setting. The pilot test sessions were organized in a class. Six students participated in the remote tensile strength labs (Fig. \ref{figure3}) over three weeks, resulting in 14 man-hours of testing.} Participants connected to TeleopLab through cellular data, home WiFi, or company WiFi. 

In the lab, a BIC staff was physically present during the sessions. The staff 
was trained to handle the preloading and unloading of samples and resolve any robot errors that occurred during the sessions. During the sessions, we recorded the time of each member's performance as part of the data collection. The cycle time was defined as the time from the start of step 2 to the next time they were about to start step 2 for the next iteration. Additionally, after each session, participants completed a user experience survey, the NASA Task Load Index (TLX) \cite{Hart1988}, which measures perceived workload, and the System Usability Scale (SUS) \cite{brooke1996sus}, which evaluates the overall usability of the system. This comprehensive data collection approach provided valuable insights into both the technical performance of the system and the user experience.

\section{RESULT AND DISCUSSION}
\subsection{Student Performance}
\begin{figure}[htbp]
      \centering
      \makebox{\includegraphics[width = 2.8in]{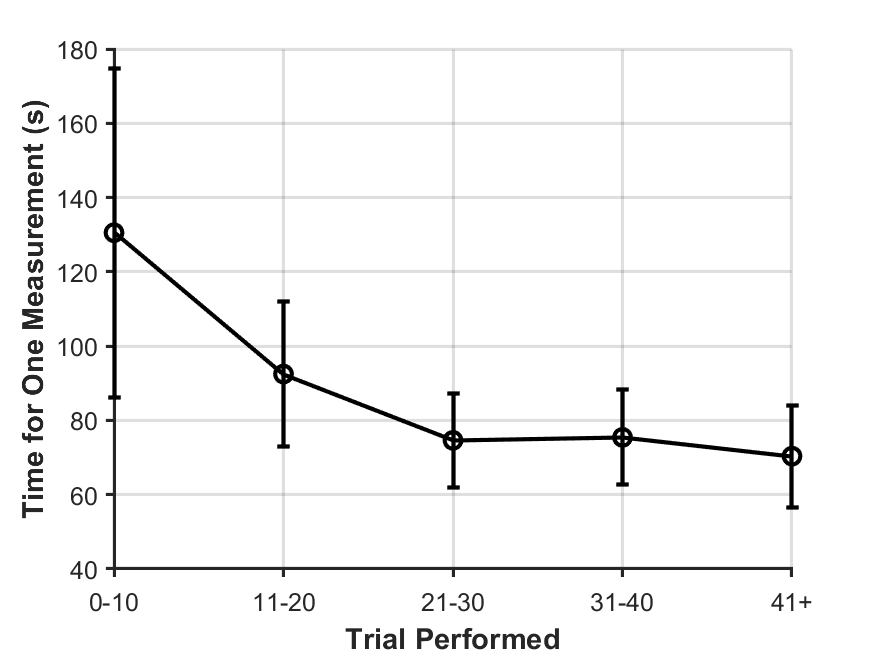}}
      \caption{\textbf{Student Performance Over Time. } The results show a decreasing trend in time as trials progress, indicating performance improvement with practice. Error bars represent the standard deviation across trials, which also decrease over time, reflecting increased consistency in performance.}
      \label{figure4}
   \end{figure}

During the initial 10 trials, the average time for one measurement was $\sim$140 seconds, but by the time students had completed 30 trials, this time decreased to $\sim$80 seconds. \YY{Fig. \ref{figure4} portrays the improvement in performance as students gain familiarity with TeleopLab.} \YY{In the early stages of the trials, some students occasionally mishandled or dropped a specimen and were not yet familiar with the perception interface, leading to higher cycle times.} By the 40th trial, the average cycle time stabilized around 60-80 seconds. The 46.1\% decrease in cycle time suggests that as students became more comfortable with the system, their efficiency improved, leading to quicker completion. 

\subsection{Quantitative User Feedback}
\begin{figure}[thpb]
      \centering
      \makebox{\includegraphics[width = 3.3in]{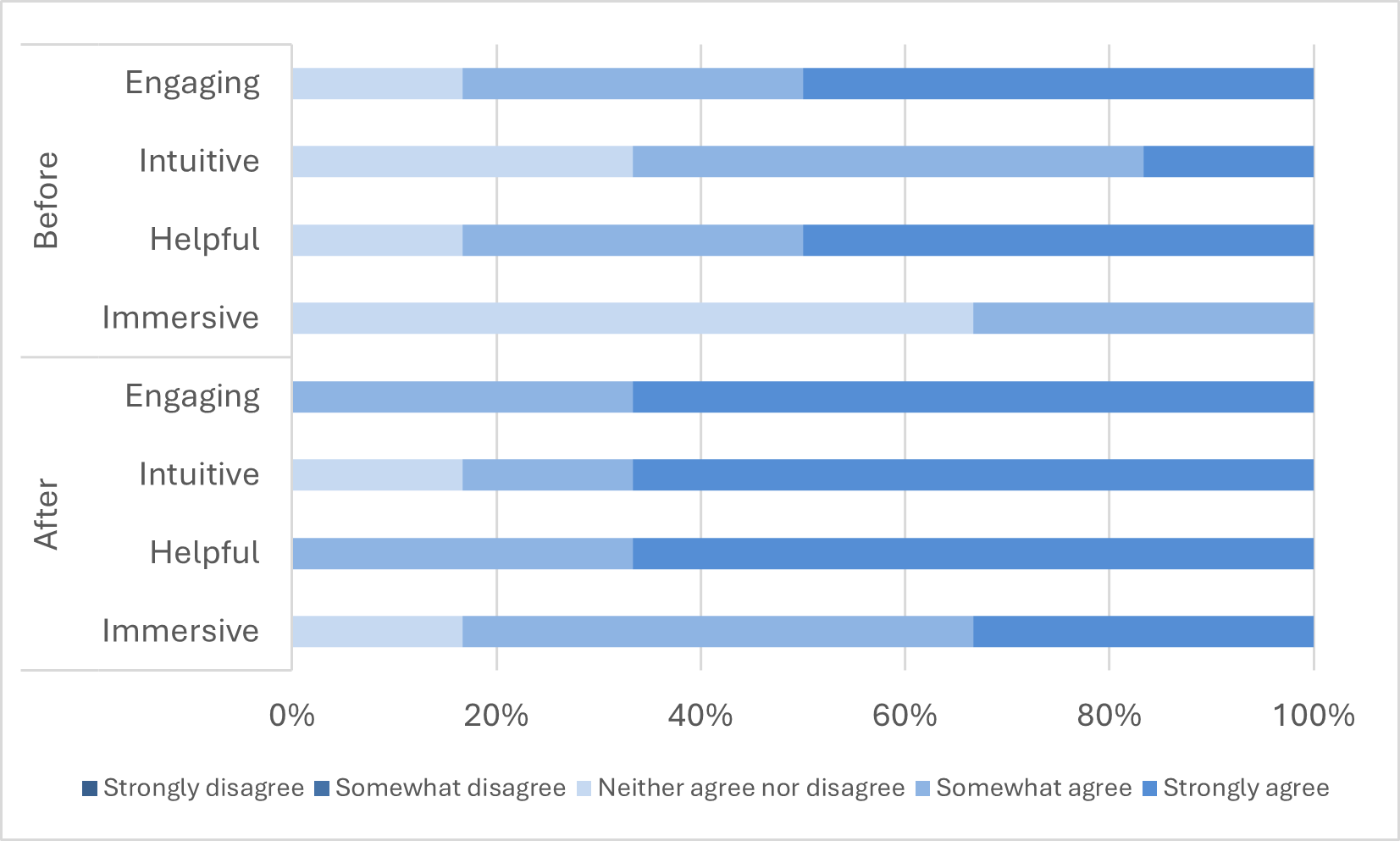}}
      \caption{\textbf{Quantitative User Feedback.} Comparison of students' expectations before using the system and their feelings after using the system across immersiveness, helpfulness in learning, intuitiveness and engagement.}
      \label{figure5}
   \end{figure}

We surveyed students' expectations of the system and how their perspectives changed after using it. The survey consisted of two sets of questions: one set asked students about their expectations before using the system, while the second captured their perceptions after the teleoperation experience. The four key categories assessed were immersiveness, helpfulness in learning, intuitiveness, and engagement. Both sets used a 5-point Likert scale, ranging from "strongly disagree" to "strongly agree." Fig. \ref{figure5} shows students' perspectives across all four aspects improved after using the system. Intuitiveness showed the most significant positive shift, indicating that the system was easy to use and aligned well with user expectations. This improvement highlights the effectiveness of the design in making the system accessible for users with minimal prior experience in teleoperation or robotics. The positive changes in immersiveness, helpfulness, and engagement further suggest that the system successfully enhanced the learning experience by providing a hands-on, engaging interaction with real equipment.

\begin{table}[h!]
\centering
\caption{NASA TLX Evaluation.}
\begin{tabular}{lcc}
\hline
\textbf{Measure}     & \textbf{Mean (1-5)} & \textbf{STD} \\
\hline
Mental               & 2.83                & 1.33         \\
Physical             & 2.17                & 1.60         \\
Temporal             & 3.00                & 1.41         \\
Effort               & 1.83                & 0.98         \\
Performance          & 3.17                & 1.17         \\
Frustration          & 2.17                & 1.17         \\
\hline
\textbf{Average (0-100)}     & \textbf{38.19}      & \textbf{20.82} \\
\hline
\end{tabular}
\label{tab:nasatlx}
\end{table}

\subsection{Data Collection}
We administered two post-use surveys: the NASA Task Load Index (NASA TLX) and the System Usability Scale (SUS). Both surveys were conducted on a 5-point Likert scale for consistency, with the NASA TLX scores later converted to a 0-100 range for analysis. NASA TLX measured participants' perceptions across six categories: mental demand, physical demand, temporal demand, performance, effort, and frustration. The overall workload score was calculated by summing these six categories, where higher scores indicate a higher perceived workload on the students. In Table \ref{tab:nasatlx}, the average workload score of 38.19 shows that while the teleoperation task posed some mental, temporal, and performance challenges, participants found the overall workload manageable, particularly in terms of physical demand, effort, and frustration.

\begin{table}
\centering
\caption{System Usability Scale (SUS) Score}
\begin{tabular}{lcc}
\hline
\textbf{SUS Questions} & \textbf{Mean (1-5)} & \textbf{STD} \\ \hline
1. Like to use frequently & 4.50 & 0.55 \\ 
2. Found system complex & 2.17 & 0.98 \\ 
3. Easy to use & 4.33 & 0.52 \\ 
4. Need technical support & 3.17 & 1.47 \\ 
5. Functions well integrated & 4.17 & 0.75 \\
6. Too much inconsistency & 2.50 & 1.38 \\ 
7. Quick to learn & 4.00 & 1.26 \\ 
8. Cumbersome to use & 2.33 & 1.03 \\ 
9. Confident using system & 4.50 & 0.55 \\ 
10. Require too much before using & 1.83 & 0.98 \\ \hline
\textbf{SUS Score (0-100)} & \textbf{73.75} & \textbf{14.21} \\ 
\hline
\end{tabular}
\label{tab:sus}
\end{table}

To complement NASA TLX, we introduced SUS to evaluate the overall usability of the system. The SUS questionnaire alternates between positive odd-numbered and negative even-numbered statements to balance feedback and minimize response bias. The SUS scores were also converted from a 5-point scale to a 0-100 range, with higher scores representing better usability. Table \ref{tab:sus} shows a score of 73.75 with a standard deviation of 14.21. A SUS score above 68 is generally considered above average, indicating that the participants found the system usable \cite{DrewMandyR.2018WDtS}. 

The high scores for frequent use (4.5), ease of use (4.33) and confidence (4.50) suggest that students were comfortable using the system. The scores for quick to learn (4.00) and requirements before using (1.83) suggest the system is intuitive. The moderate scores for the need for technical support (3.17) and inconsistency (2.5) indicate occasional issues during the experience, mainly caused by the connection drop, packet loss, or drifting of the coordinate frame on the phone. 

\section{Conclusion}
In this study, we introduced TeleopLab, a phone-based teleoperation system to provide hands-on learning in a remote setting. The system bridges the gap between traditional labs and remote education by enabling real-time control of a robotic arm and lab equipment. Our results showed a promising learning curve, with cycle times improving by over 40\% as students became more familiar with the system. Three quantitative surveys—capturing students' perspectives before and after use, along with NASA TLX and SUS—demonstrated overall improvements in students' perceptions. TeleopLab successfully preserved key aspects of hands-on learning, offering scalability and flexibility for future applications in remote STEM education. \YY{Our future work will focus on advancing the system to enable high-precision teleoperation for highly dexterous tasks. We intend to integrate an adaptive feedback mechanism to enhance situational awareness.} \ZC{Additionally, we aim to enhance the perception of our system through more intuitive interfaces.} \YY{Our goal is to make TeleopLab a promising tool for future applications in STEM education and beyond.}
\newpage




\bibliography{export}

\end{document}